\documentclass[runningheads]{llncs}
\usepackage[T1]{fontenc}
\usepackage{graphicx}
\usepackage{amsmath}
\usepackage{mathtools}
\usepackage{amsmath}
\usepackage{amssymb,amsfonts}
\usepackage{algcompatible}

\usepackage{amsfonts} 
\usepackage{multirow}
\usepackage{multicol}
\usepackage{booktabs}
\usepackage{multirow}
\usepackage{multicol}
\usepackage{adjustbox}
\usepackage{amsmath}

\usepackage{enumitem}
\usepackage{booktabs}
\usepackage{siunitx}
\usepackage{xcolor} 

\begin{document}
\title{
CoBooM: Codebook Guided Bootstrapping for Medical Image  Representation Learning  }
\titlerunning{CoBooM}
%
\author{Azad Singh\orcidID{0000-0002-6607-1130} \and
Deepak Mishra \orcidID{0000-0002-4078-9400} }

%
\authorrunning{A. Singh \and D. Mishra}
%
\institute{Indian Institute of Technology, Jodhpur 342307, Rajasthan, India 
\email{\{singh.63,dmishra\}iitj.ac.in}}
\maketitle              
\begin{abstract}
Self-supervised learning (SSL) has emerged as a promising paradigm for medical image analysis by harnessing unannotated data. Despite their potential, the existing SSL approaches overlook the high anatomical similarity inherent in medical images. This makes it challenging for SSL methods to capture diverse semantic content in medical images consistently. This work introduces a novel and generalized solution that implicitly exploits anatomical similarities by integrating codebooks in SSL. The codebook serves as a concise and informative dictionary of visual patterns, which not only aids in capturing nuanced anatomical details but also facilitates the creation of robust and generalized feature representations. In this context, we propose \textit{CoBooM}, a novel framework for self-supervised medical image learning by integrating continuous and discrete representations. The continuous component ensures the preservation of fine-grained details, while the discrete aspect facilitates coarse-grained feature extraction through the structured embedding space. To understand the effectiveness of CoBooM, we conduct a comprehensive evaluation of various medical datasets encompassing chest X-rays and fundus images. The experimental results reveal a significant performance gain in classification and segmentation tasks.

\end{abstract}

\section{Introduction}
Expensive annotations for medical images promote Self-Supervised Learning (SSL)~\cite{simclr,moco,byol,simsiam}. 
Recent developments demonstrate its effectiveness across diverse modalities, such as X-rays, MRIs, CT, and histopathology~\cite{huang2023self,caid}. However, despite advancements, existing methods like SimCLR~\cite{simclr}, MoCo~\cite{moco}, BYOL~\cite{byol}, and VICReg~\cite{vicreg} encounter challenges when applied to medical images, in terms of effectively creating positive and negative pairs. 
The complexity occurs due to inherent feature overlapping among different anatomical sub-structures and across diverse image samples. Current SSL methods oversee the anatomical overlapping and, thus, potentially compromise the model's performance and generalization capabilities. 

In this work, we propose a simple yet effective technique involving learning generalized features guided by a codebook~\cite{vqvae,zheng2023online}, enabling the capturing of concise discrete features. By associating similar anatomical features with common codes and distinguishing features with distinct codes, the codebook facilitates a structured learning process, which overcomes the challenges associated, such as defining effective positive and negative pairs~\cite{wang2022contrastive}. This establishes a systematic representation where recurring patterns are encoded consistently. For instance, the presence of lung fields, ribs, and cardiac contours, common across chest X-rays, may share the same or similar codes, providing a concise and shared representation of prevalent features and creating a sparse but informative summary of the entire dataset. 
This introduces a strong structured inductive bias by implicitly guiding the SSL model toward making assumptions about the common patterns and structures present.

In this context, we propose an SSL framework named CoBooM: Codebook Guided Bootstrapping for Medical Image Representation Learning. Specifically, CoBooM encompasses a Context and Target Encoders for learning continuous features and a Quantizer module to quantize the features using codebook and integrate them with continuous features using the novel DiversiFuse sub-module. The DiversiFuse sub-module utilizes cross-attention mechanisms that capitalize on the complementary information offered by these two representations. The introduction of the codebook encourages the SSL model to recognize and prioritize the shared generalized common features during the training process. In addition, the complementary integration of the continuous and discrete representations allows the model to capture fine-grained features, contributing to a smooth and rich embedding space. This leads to a more holistic and refined understanding of the underlying data.
We conduct experiments across diverse modalities to validate its effectiveness, encompassing chest X-ray and fundus images. We evaluate the proposed approach under linear probing and semi-supervised evaluation protocols and observe more than 3\% performance gains in downstream classification and segmentation tasks. 

\section{Background}
\textbf{Discriminative SSL Approaches}:
Discriminative SSL has seen advancements with approaches like SimCLR~\cite{simclr}, MoCo~\cite{moco,mocov2}, BYOL~\cite{byol}, Barlow-Twins~\cite{barlow}, that captures generalized features by enhancing the similarity between positive pairs while maximizing the dissimilarity between negative pairs either explicitly or implicitly. In the domain of medical images, discriminative SSL techniques, especially contrastive approaches, have gained substantial attention and found meaningful applicability. Various adaptations of contrastive methods, like MoCo-CXR~\cite{sowrirajan2021mococxr}, for chest X-rays, MICLe~\cite{azizi2021big} using multiple patient images, and MedAug~\cite{vu2021medaug} with metadata-based positive pair selection, contribute to the improvement of medical image representations. Simultaneously, another approach, DiRA~\cite{dira}, unites the discriminative, restorative, and adversarial learning to capture the complementary features.
Zhou \textit{et al.} propose PCRL~\cite{pcrlv1} for X-ray and CT modalities, later improved with PCRLv2~\cite{pcrlv2} addressing pixel-level restoration and scale information. Kaku \textit{et al.} enhance contrastive learning with intermediate-layer closeness in their approach~\cite{intermoco}. In \cite{histo1}, SimCLR was used for pre-training on multiple unlabeled histopathology datasets, improving feature quality and superior performance over ImageNet-pretrained networks. In other studies~\cite {histo2,histo3}, authors showcased the efficacy of different SSL methods on large-scale pathology data. While the existing methods show advancements, however they oversight the significant anatomical similarities in medical data. The proposed approach implicitly harnesses the anatomical similarities to capture more informative features.

\textbf{Codebook in Medical Image Analysis:
}Using codebook in medical image analysis holds the promising potential 
~\cite{cb4}. By discretizing the data, codebooks can simplify complex medical image features, making them easier to analyze~\cite{cb1,cb2}. Recent studies~\cite{cb5,cb8} highlight the effectiveness of learning discrete representations through codebooks across various domains in achieving interpretable and robust medical image retrieval, generation, recognition, and segmentation.

\begin{figure*}[t]
\centering
\includegraphics[width=1.0\textwidth]{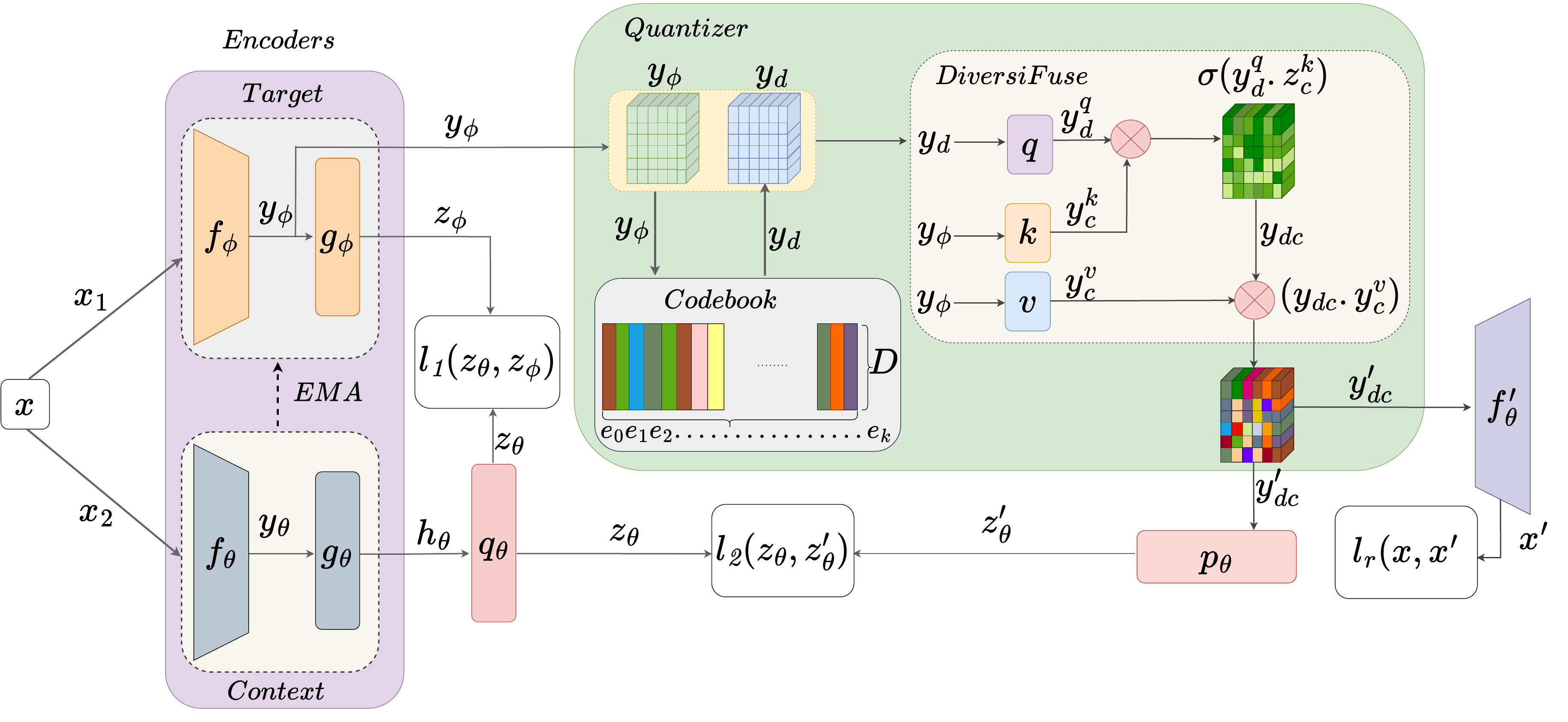}
\caption{The architecture overview of the proposed framework. EMA is an exponential moving average used to update the parameters of the Target encoder. $g_\theta$ and $g_\phi$ are the three MLP networks that serve as projection heads for Context and Target encoders. $f'_\theta$ serves as the decoder network.}
\label{main_fig}
\end{figure*}

\section{Methodology}
Fig.~\ref{main_fig}
provides an architectural layout of the proposed SSL framework, comprising a Context encoder parameterized by $\theta$, 
a Target encoder parameterized by $\phi$ and a Quantizer module. Additionally, two projection heads are denoted as $q_\theta$ and $p_\theta$ and a decoder $f'_\theta$. The proposed framework adheres to the self-distillation-based non-contrastive SSL paradigm~\cite{byol}. The parameters $\theta$ undergo updates through back-propagation of the loss, while the parameters $\phi$ are the earlier version of the $\theta$, updated using exponential moving average(EMA). Given an input sample $x$, it creates two augmented views $x_1$ and $x_2$
by applying the random set of augmentations. $x_1$ is processed by $f_\phi$ to output feature map $y_\phi$ while $f_\theta$ produces $y_\theta$ from $x_2$. Further, $y_\theta$ and $y_\phi$ after passing through the global average pooling layer, fed to predictor heads $g_\theta$ and $g_\phi$ to output the embeddings $z_\theta$ and $z_\phi$ carrying the global features. Subsequently, the target feature map $y_\phi$ is quantized through the \textit{Quantizer} module, utilizing a \textit{Codebook} and \textit{DiversiFuse} module to represent and compress the features effectively. The following subsection provides details of the proposed quantization process.

\subsection{Quantizer}
The Quantizer module utilizes codebook, a predefined table containing $K$ discrete codewords represented as vectors $e_k$, each of size $D$. These codewords are employed to quantize 
the lower-dimensional continuous feature maps $y_\phi$ received from the target encoder $f_\phi$. The Quantizer module compares the features from $y_\phi$ with each $K$ codewords in the codebook to measure similarity by employing the Euclidean distance. 
The module identifies the closest codeword to the encoded data through an iterative process across the codebook. Subsequently, the module replaces the continuous encoded data $y_\phi$ with the selected codewords, effectively transforming the representation from continuous to discrete $y_d$. This quantization is executed with the objective of minimizing the quantization loss $\mathcal{L}_q= l_{cb} + \alpha* l_{ce}$ comprising of two terms, codebook loss ($l_{cb} =  ||SG[y_\phi] - e_k||^2_2$) and the commitment loss ($l_{ce} = ||y_\phi - SG[e_k]||^2_2$). Here $SG$ denotes the stop-gradient operator and $\alpha$ specifies the weight of $l_{ce}$. The codebook loss guides the adjustment of the codewords $e_k$ towards $y_\phi$. Simultaneously, the commitment loss enforces $y_\phi$ to adhere to specific embeddings in the codebook, thus preventing unregulated expansion.

\subsubsection{DiversiFuse (Feature Fusion with Multi-Head Cross Attention):}
Within Quantizer, the DiversiFuse sub-module guides the model through discrete representations $y_d$ in determining which parts of the continuous information $y_\phi$ are more relevant. It enables the model to learn to focus on different aspects of the continuous representation based on the specific values in the discrete features, potentially capturing more complex patterns and dependencies within the data. It involves a multi-head cross-attention mechanism where the quantized features $y_d$ pass through $q$ to output $y^q_d$, and the continuous features $y_\phi$ pass through $k$ and $v$ to output $z^k_c$ and $y^v_c$ respectively. The similarity scores between discrete queries $y^q_d$ and the continuous keys $y^k_c$ are calculated as $S_{Score}(y^q_d, y^k_c) = z^q_d \cdot {y^k_c}^T$. Subsequently, the scores are transformed into attention weights using the softmax function: $\sigma(S_{Score}(y^q_d, y^k_c))$ denoted as $y_{dc}$. The continuous values $y^v_c$ are then weighted by the attention weights $y_{dc}$ and summed: $W_{Sum}(y_{dc},y^v_c ) = \sum y_{dc} \cdot y^v_c $. The keys $y^k_c$ help determine which parts of the continuous information should be attended to, and the values provide the actual information to be attended to. The process is repeated for all attention heads. The resulting aggregated representation $y'_{dc}$ is obtained through concatenation across all attention heads. This integration of discrete and continuous representations enables the exchange of complementary information, enhancing the model's ability to capture complex patterns and improve performance.

\subsection{Loss Function}
The output of the Quantizer module, denoted as $y'_{dc}$, undergoes an average pooling layer and is subsequently projected into a lower-dimensional space using the projection head $p_\theta$. The resulting output of $p_\theta$ is denoted as $z'_\theta$. To optimize the parameters $\theta$, the similarity scores between $z_\theta$ and $z_\phi$, as well as between $z_\theta$ and $z'_\theta$, are calculated using the loss function defined in Equation~\eqref{loss1}.
\begin{equation}
    \mathcal{L}_1 =  \frac{\langle z_\theta , z_\phi \rangle}{||z_\theta||_2 . ||z_\phi||_2},  \mathcal{L}_2 =  \frac{\langle z_\theta , z'_\theta \rangle}{||z_\theta||_2 . ||z'_\theta||_2}
    \label{loss1}
\end{equation}
Additionally, $y'_{dc}$ also fed to the decoder $f'_\theta$ to output the reconstructed image $x'$, enabling the model to capture local complementary features, formulated as $\mathcal{L}_{r} = ||x - x'||_2$. 
The final loss $L_\theta = \alpha(\mathcal{L}_1 + \mathcal{L}_2) + \mathcal{L}_q + \gamma\mathcal{L}_{r} $, where $\alpha$ and $\gamma$ set to 0.5. Additionally, the symmetric form of the loss $L_\theta$
is utilized by interchangeably feeding the views $x_1$ and $x_2$ to $f_\theta$ and $f_\phi$.

\section{Experimental Setup }
\textbf{Descriptions of Datasets:}
For pre-training, we utilize a publicly available official train set from NIH-Chest X-ray 14~\cite{wang2017chestx} consisting of 86,524 X-ray images and the fundus images from the EyePACS~\cite{eyepacks} dataset. The downstream classification task is performed on the officially available test set, with 25,596 samples and the retinal images from MuReD~\cite{mured} and ODIR~\cite{odir1,odir2} datasets. To assess the performance for the downstream segmentation task, we utilize the SIIM-ACR~\cite{siim} dataset for pneumothorax detection.\\
\textbf{Implementation Details:}
We train the models on the Nvidia RTX A6000 with the PyTorch framework. For backbone encoders ($f_\theta$ and $f_\phi$), we use ResNet18 architecture, with an input image size of 224$\times$224, batch size of 64, and number of epochs of 300. The number of codebook vectors are 1024, each of size 512. All projection and prediction heads are three-layer MLP networks with an output size 256. For optimizing the parameters $\theta$, we employ LARS \cite{you2017large} optimization, a base learning rate set at $0.02$. Additionally, we implement a cosine decay learning rate scheduler without restarts.\\
\textbf{Baselines for Comparison:}
To assess the performance of our proposed approach, we compare it with supervised learning, with random initialization (Sup.) and several established SSL methods, encompassing contrastive, non-contrastive, and clustering-based techniques including SimCLR~\cite{simclr}, BYOL~\cite{byol}, VICReg~\cite{vicreg}, SwAV~\cite{SwaV}, DiRA~\cite{dira}, CAiD~\cite{caid} and PCRLv2~\cite{pcrlv2}.
Notably, we conduct the pre-training for the baselines following their official implementations and using the same training protocol as our proposed method. 

\section{Results and Discussion}

          


\begin{table*}[htbp]
    \setlength{\tabcolsep}{3pt}
    \centering
    \caption{Performance evaluation of the proposed approach in terms of AUC score on the NIH, MuRed, and the ODIR datasets, and dice score for the pneumothorax segmentation (SIIM) under linear probing. The best results are bold,  SD is not shown due to low variability.}
    \label{tab:linerprobe}
    \begin{tabular}{ccccccc|cc}
        \toprule
        \multirow{2}{*}{\textbf{Methods}}&\multicolumn{5}{c}{\textbf{NIH}} & \multicolumn{1}{c}{\textbf{SIIM}}& {\textbf{MuReD}}& \textbf{ODIR}  \\
        \cmidrule(lr){2-6} \cmidrule(lr){7-7} \cmidrule(lr){8-8} \cmidrule(lr){9-9}
        & {1\%} & {5\%} & {10\%}   & {30\%} &{All} &{All} & 10\% & 10\% \\
        \midrule
        Sup.  & 51.6 & 55.1 & 57.1 & 61.1 & 61.8 &48.4 & 58.6 & 56.4\\ 
        SimCLR  & 56.9 & 59.7 & 62.7 & 67.6 & 70.0  &50.3& 72.1 & 70.2\\
        BYOL  & 54.7 & 58.3 & 61.7 & 66.3  & 69.0  &49.8& 70.5& 67.4\\
        SwAV  & 55.5 & 59.1 & 62.4 & 67.7  & 70.2  &53.4& 71.6&70.8\\
        VICReg  & 58.7 & 60.7 & 62.7 & 66.2  & 67.3 & 48.7&72.4&66.5 \\
        CAiD  & 63.7 & 67.2 & 68.9 & 70.3  & 73.5& 55.3&70.7 & 69.5\\
        PCRLv2 & 61.9 & 66.4 & 68.3 & 71.5  & 73.8&56.4&72.6 & 72.4 \\
        DiRA & 60.8 & 65.8 & 68.6 & 72.6  & 74.1&56.8&71.7&70.8 \\
        \midrule
        Ours w/o Dec. & \textbf{65.1} & \textbf{70.1} & 72.0 & \textbf{73.6}  & \textbf{74.8}& 55.6 & 75.8 & \textbf{76.0} \\

        Ours w/ Dec. & 64.9 &  70.3 & \textbf{72.4} & 73.3 & 74.3  & \textbf{57.5} & \textbf{76.0} & 75.3\\
                
        Ours w/o DF. & 63.3 & 68.6 & 70.9 & 72.1 & 73.4 & 54.9 & 74.6 & 73.8 \\

        \bottomrule
    \end{tabular}
\end{table*}

\begin{figure}[t!]
\setlength\tabcolsep{0.17pt}
\centering
\begin{tabular}{ccc}
&  \hspace{-0.4cm}Ours \hspace{0.2cm} DiRA \hspace{0.3cm}  PCRLv2 \hspace{0.1cm}  CAiD &  \hspace{-0.2cm}Ours \hspace{0.3cm}  DiRA \hspace{0.3cm}  PCRLv2 \hspace{0.1cm}  CAiD  \\

\rotatebox{90}{Atelec.} \hspace{0.1cm}&\includegraphics[width=.45\linewidth]{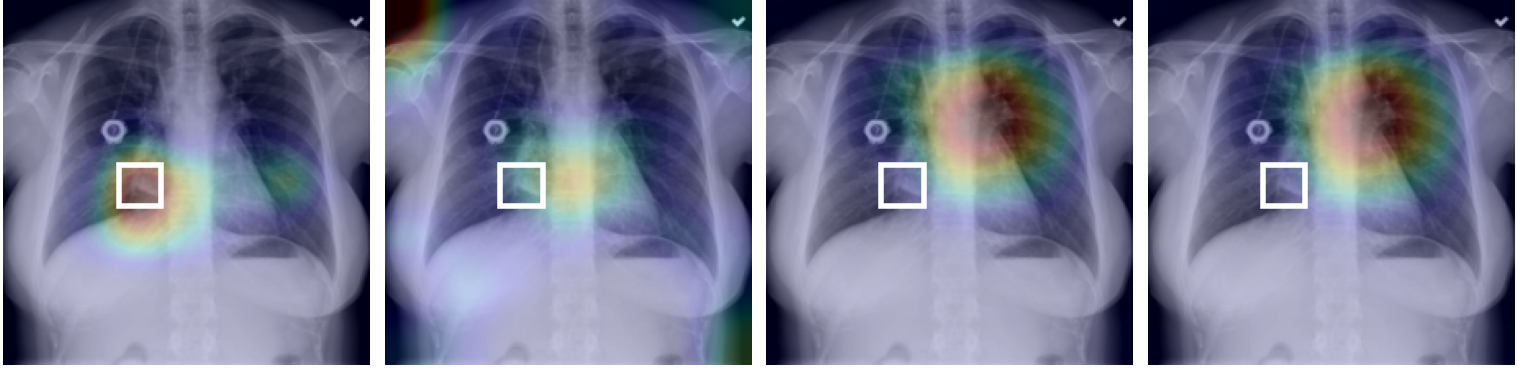} 
\rotatebox{90}{Cardio.}\hspace{0.1cm}&\includegraphics[width=.45\linewidth]{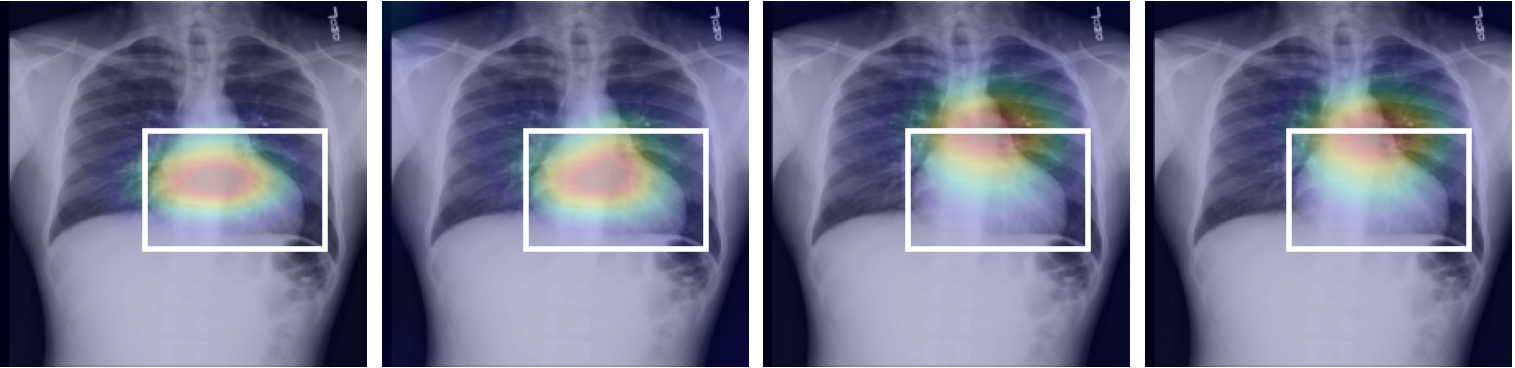}\\

\rotatebox{90}{Effus.}\hspace{0.1cm}&\hspace{0.1cm}\includegraphics[width=.45\linewidth]{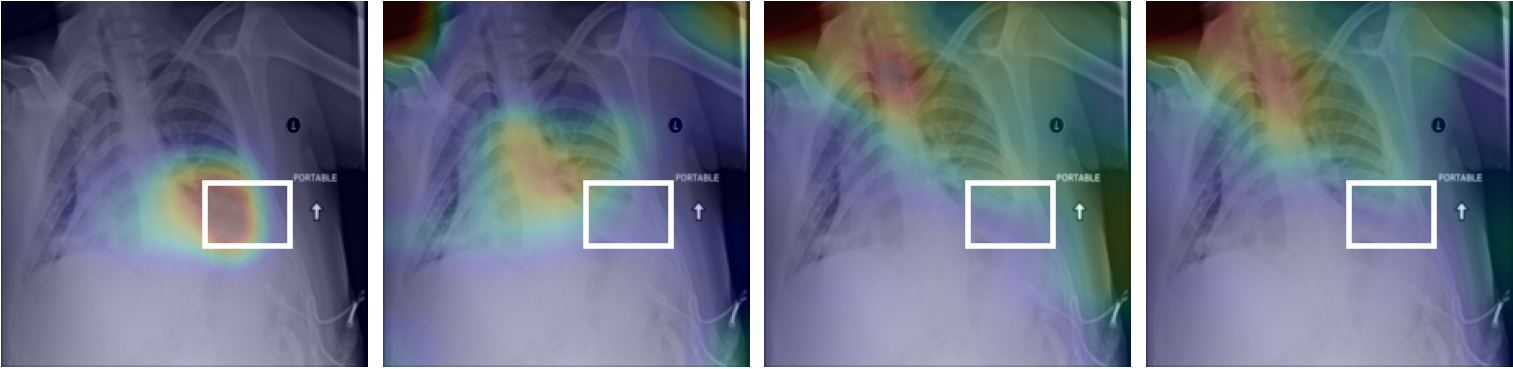} 
\rotatebox{90}{Mass}\hspace{0.1cm}&\hspace{0.1cm}\includegraphics[width=.45\linewidth]{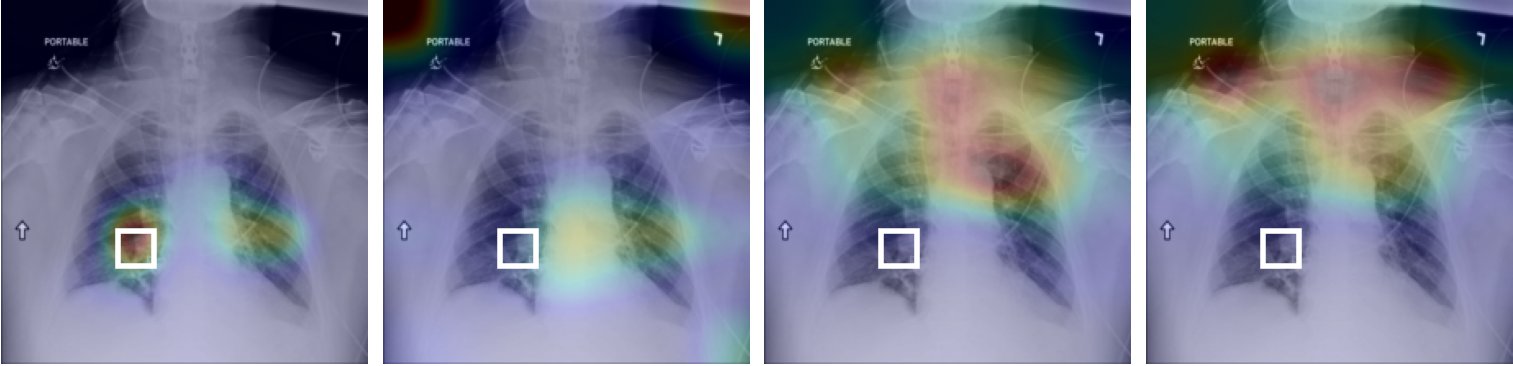} \\

\end{tabular}
\caption[short]{Diagnostic maps for Atelectasis, Effusion, Cardiomegaly, and Mass corresponding to the X-ray images from NIH indicate that CoBooM captures pathological features effectively compared to other best-performing baseline methods. The bounding box indicates the ground truth. 
\label{fig_cam: gradcam}}
\end{figure}

\textbf{Linear Probing Evaluation:} Table~\ref{tab:linerprobe} presents the experimental results on NIH and SIIM-ACR datasets under linear probing protocol. Specifically, the parameters of encoder $f_\theta$ remain frozen while that of the linear layer get updated. For NIH, we evaluate the performance by sample labeled subsets from the official train set and report the official test set results in terms of AUC score. Similarly, on MuRed and ODIR datasets, the test set AUC score is reported by evaluating 10\% of labeled training data. For pneumothorax segmentation on SIIM-ACR, we report the results in terms of dice score by updating the parameters of the decoder network while that of the encoder remains frozen. Supervised learning (Sup.) notably yields lower AUC scores than the SSL methods. The proposed approach consistently outperforms other baselines across varying degrees of labeled data, specifically for the 1\%  subset from NIH, the our approach achieves the highest AUC score of 65.1\% with an average performance gain of more than \textbf{3\%} from all the baseline methods. Fig~\ref{fig_cam: gradcam} presents the diagnostic maps for different pathological conditions, corresponding to the 10\% labeled samples from NIH. A similar trend is observed for MuReD and the ODIR dataset, where the proposed approach outperforms the baselines with a considerable average margin of more than \textbf{3\%}. This indicates the method's ability to extract meaningful representations from unlabeled data for the subsequent downstream training using limited labeled samples. Furthermore, a similar improvement in AUC scores is observed with increased labeled data. The proposed approach also results in the highest dice score of 57.5\%  on pneumothorax segmentation, with an improvement of 1\% compared to the best-performing baseline. 


\begin{table*}[htbp]
\setlength{\tabcolsep}{3pt}
\centering
\caption{Semi-supervised fine-tuning evaluation in terms of AUC score (\%) on the NIH, MuRed, and the ODIR datasets, and dice score for the pneumothorax segmentation. }
\label{tab:semi}
\begin{tabular}{ccccccc|cc}
        \toprule
\multirow{2}{*}{\textbf{Methods}}&\multicolumn{5}{c}{\textbf{NIH}} & \multicolumn{1}{c}{\textbf{SIIM}}& {\textbf{MuReD}}& \textbf{ODIR}  \\
        \cmidrule(lr){2-6} \cmidrule(lr){7-7} \cmidrule(lr){8-8} \cmidrule(lr){9-9}
        & {1\%} & {5\%} & {10\%}   & {30\%} &{All} &{All} & 10\% & 10\% \\
\midrule
SUP. & 57.7 & 62.7 & 65.6 & 70.7 & 74.1 &51.2&66.7 &63.2\\
SimCLR & 62.1 &65.7& 68.9 & 72.2  & 75.6 &53.3&80.9 &73.4 \\
BYOL &61.0 & 65.2& 67.7 & 71.6 & 74.8&  52.8&78.6 &71.3\\
SwAV  & 61.7 & 65.6 & 66.9 & 72.1  & 75.8  &54.4&79.4 &72.7\\
VICReg & 60.0&64.8 & 68.4 & 71.8 & 75.4 & 54.4&78.3 &72.9\\
CAiD & 64.4 &69.6& 71.3 & 73.8 & 77.4 & 56.5 &81.0 &73.1\\
PCRLv2 & 63.0 &68.7& 70.6 & 73.1 & 76.1& 57.3&82.4 &74.6\\
DiRA & 62.7 & 67.3 & 71.2 & 74.5  & 77.8&58.8&81.6 &73.4 \\
\midrule
Ours w/o Dec. & \textbf{65.8} & \textbf{70.6} & \textbf{72.3} & \textbf{76.7} & \textbf{79.6}& 57.8 &84.4 &\textbf{75.8} \\
Ours w/ Dec. & 65.6 &  70.8 & 72.1 & 77.1 & 79.3 & \textbf{59.6}& \textbf{84.8}&75.7\\
Ours w/o DF. & 63.7 & 70.0 & 72.2 & 76.3 & 78.9 &57.1& 83.1 &74.2\\

\bottomrule
\end{tabular}
\end{table*}
\noindent \textbf{Semi-Supervised Evaluation:}
Table~\ref{tab:semi} presents the test set performance of the baseline methods and the proposed approach under semi-supervised evaluation on the NIH, SIIM-ACR, MuRed, and ODIR datasets,  where we fine-tune the parameters of backbone encoder $f_\theta$ also along with the linear layer. We present the official test set performance evaluation in terms of AUC score on the NIH, MuReD, and ODIR by fine-tuning the model using various subsets of labeled data extracted from the training samples. We observe consistently superior performance of the proposed approach over existing SSL methods across all the subsets. Notably, our method achieves the highest AUC score of 65.8\%, with 1\% of the training samples surpassing the baselines by a margin exceeding 2\%. The trend persists as the labeled data increases to 100\%, with the proposed approach consistently outperforming the baselines and maintaining an average gain of 2\%. MuRed and ODIR datasets have a similar performance gain, with the highest AUC scores of 84.8 and 75.8, respectively. For pneumothorax segmentation also, we observe the highest dice score of 59.6\% with a margin of more than 2\% compared to the best-performing baseline method.\\
\textbf{Optimal Performance with Minimal Fine-Tuning:}
Upon comparing the results presented in Table~\ref{tab:linerprobe} and~\ref{tab:semi}, a noteworthy observation is that our proposed method demonstrates minimal or no need for fine-tuning of the backbone encoder, especially with lower numbers of labeled training samples. Specifically, at 1\%, the proposed method achieves AUC scores of 65.1\% and 65.8\% under the linear-probing and semi-supervised fine-tuning evaluation protocols, respectively. Similarly, for 5\% and 10\% labeled training samples, our method's AUC scores remain comparable with negligible margins. This trend contrasts baseline methods, where a substantial performance gain is observed from linear probing to semi-supervised fine-tuning. 
This highlights the effectiveness of our proposed method while demonstrating a remarkable capacity to achieve optimal performance with minimal fine-tuning to adapt to different tasks. This signifies the proposed approach's adaptability and highlights its potential to derive meaningful and transferable representations with minimal fine-tuning, which aligns with the practical requirements of real-world settings where computational resources may be limited.\\

\noindent \textbf{Ablation Studies:}
We conduct an ablation study to examine the impact of different components of the proposed approach under both linear probing and semi-supervised evaluation protocols. In our first study, we evaluate the model's performance by performing the pre-training, with and without the decoder by keeping the DiversiFuse module.  We pre-train the model without the DiversiFuse module and the decoder for another study. Table~\ref{tab:linerprobe} and~\ref{tab:semi} present the test set results across various downstream tasks for these studies. We observe no effect of the decoder on the model's performance during classification tasks in the downstream evaluations. However, while evaluating the performance on the segmentation task, we observed superior performance when pre-training the model with the decoder under both evaluation protocols. When pre-train the model without the DiversiFuse sub-module in the Quantizer, we observe a decline of around 2\% across all tasks  
on evaluating the model's performance under linear probing. Under semi-supervised evaluation, the model can maintain its performance even without the DiversiFuse sub-module, however, for classification with 1\% labeled samples from NIH, we observe a degradation in AUC score of 2\%. This highlights the importance of the DiversiFuse sub-module in improving the quality of the learned representations with the help of discrete features.  

\section{Conclusion} 

In this work, we propose an efficient SSL pre-training by integrating the discrete and continuous features with the help of a codebook. We propose a novel DiversiFuse sub-module, which guides the model in learning generalized and better representation and does not require much fine-tuning, especially when labeled data is limited. We highlight the proposed model's ability to capture complex medical attributes with limited resource availability through empirical studies. We evaluate the performance of the proposed approach by comparing it with various SSL methods under both linear probing and semi-supervised evaluations for both classification and segmentation tasks. This highlights its effectiveness in handling various tasks associated with medical image analysis.
\bibliographystyle{splncs04}
\bibliography{ref}

\begin{thebibliography}{10}
\providecommand{\url}[1]{\texttt{#1}}
\providecommand{\urlprefix}{URL }
\providecommand{\doi}[1]{https://doi.org/#1}

\bibitem{siim}
Society for imaging informatics in medicine: Siim-acr pneumothorax segmentation (2019), \url{https://www.kaggle.com/c/siim-acr-pneumothorax-segmentation/overview/description}

\bibitem{azizi2021big}
Azizi, S., Mustafa, B., Ryan, F., Beaver, Z., Freyberg, J., Deaton, J., Loh, A., Karthikesalingam, A., Kornblith, S., Chen, T., Natarajan, V., Norouzi, M.: Big self-supervised models advance medical image classification (2021)

\bibitem{vicreg}
Bardes, A., Ponce, J., LeCun, Y.: Vicreg: Variance-invariance-covariance regularization for self-supervised learning (2022)

\bibitem{histo3}
Boyd, J., Liashuha, M., Deutsch, E., Paragios, N., Christodoulidis, S., Vakalopoulou, M.: Self-supervised representation learning using visual field expansion on digital pathology. In: Proceedings of the IEEE/CVF International Conference on Computer Vision. pp. 639--647 (2021)

\bibitem{SwaV}
Caron, M., Misra, I., Mairal, J., Goyal, P., Bojanowski, P., Joulin, A.: Unsupervised learning of visual features by contrasting cluster assignments. Advances in Neural Information Processing Systems  \textbf{33},  9912--9924 (2020)

\bibitem{simclr}
Chen, T., Kornblith, S., Norouzi, M., Hinton, G.: A simple framework for contrastive learning of visual representations (2020)

\bibitem{mocov2}
Chen, X., Fan, H., Girshick, R., He, K.: Improved baselines with momentum contrastive learning. arXiv preprint arXiv:2003.04297  (2020)

\bibitem{simsiam}
Chen, X., He, K.: Exploring simple siamese representation learning (2020)

\bibitem{histo1}
Ciga, O., Xu, T., Martel, A.L.: Self supervised contrastive learning for digital histopathology. Machine Learning with Applications  \textbf{7},  100198 (2022)

\bibitem{eyepacks}
Dugas, E., Jared, Jorge, Cukierski, W.: Diabetic retinopathy detection (2015), \url{https://kaggle.com/competitions/diabetic-retinopathy-detection}

\bibitem{cb5}
Gangloff, H., Pham, M.T., Courtrai, L., Lef{\`e}vre, S.: Leveraging vector-quantized variational autoencoder inner metrics for anomaly detection. In: 2022 26th International Conference on Pattern Recognition (ICPR). pp. 435--441. IEEE (2022)

\bibitem{cb4}
Gorade, V., Mittal, S., Jha, D., Bagci, U.: Synergynet: Bridging the gap between discrete and continuous representations for precise medical image segmentation. In: Proceedings of the IEEE/CVF Winter Conference on Applications of Computer Vision. pp. 7768--7777 (2024)

\bibitem{byol}
Grill, J.B., Strub, F., Altché, F., Tallec, C., Richemond, P.H., Buchatskaya, E., Doersch, C., Pires, B.A., Guo, Z.D., Azar, M.G., Piot, B., Kavukcuoglu, K., Munos, R., Valko, M.: Bootstrap your own latent: A new approach to self-supervised learning (2020)

\bibitem{dira}
Haghighi, F., Taher, M.R.H., Gotway, M.B., Liang, J.: Dira: Discriminative, restorative, and adversarial learning for self-supervised medical image analysis. In: Proceedings of the IEEE/CVF Conference on Computer Vision and Pattern Recognition. pp. 20824--20834 (2022)

\bibitem{moco}
He, K., Fan, H., Wu, Y., Xie, S., Girshick, R.: Momentum contrast for unsupervised visual representation learning (2020)

\bibitem{huang2023self}
Huang, S.C., Pareek, A., Jensen, M., Lungren, M.P., Yeung, S., Chaudhari, A.S.: Self-supervised learning for medical image classification: a systematic review and implementation guidelines. NPJ Digital Medicine  \textbf{6}(1), ~74 (2023)

\bibitem{odir2}
kaggle: Ocular disease recognition, \url{https://www.kaggle.com/andrewmvd/ocular-disease-recognition-odir5k}

\bibitem{intermoco}
Kaku, A., Upadhya, S., Razavian, N.: Intermediate layers matter in momentum contrastive self supervised learning. Advances in Neural Information Processing Systems  \textbf{34},  24063--24074 (2021)

\bibitem{histo2}
Kang, M., Song, H., Park, S., Yoo, D., Pereira, S.: Benchmarking self-supervised learning on diverse pathology datasets. In: Proceedings of the IEEE/CVF Conference on Computer Vision and Pattern Recognition. pp. 3344--3354 (2023)

\bibitem{cb1}
Kobayashi, K., Hataya, R., Kurose, Y., Miyake, M., Takahashi, M., Nakagawa, A., Harada, T., Hamamoto, R.: Decomposing normal and abnormal features of medical images for content-based image retrieval of glioma imaging. Medical image analysis  \textbf{74},  102227 (2021)

\bibitem{mured}
Rodr{\'\i}guez, M.A., AlMarzouqi, H., Liatsis, P.: Multi-label retinal disease classification using transformers. IEEE Journal of Biomedical and Health Informatics  (2022)

\bibitem{sowrirajan2021mococxr}
Sowrirajan, H., Yang, J., Ng, A.Y., Rajpurkar, P.: Moco-cxr: Moco pretraining improves representation and transferability of chest x-ray models (2021)

\bibitem{caid}
Taher, M.R.H., Haghighi, F., Gotway, M.B., Liang, J.: Caid: Context-aware instance discrimination for self-supervised learning in medical imaging. In: International Conference on Medical Imaging with Deep Learning. pp. 535--551. PMLR (2022)

\bibitem{vqvae}
Van Den~Oord, A., Vinyals, O., et~al.: Neural discrete representation learning. Advances in neural information processing systems  \textbf{30} (2017)

\bibitem{vu2021medaug}
Vu, Y.N.T., Wang, R., Balachandar, N., Liu, C., Ng, A.Y., Rajpurkar, P.: Medaug: Contrastive learning leveraging patient metadata improves representations for chest x-ray interpretation (2021)

\bibitem{cb2}
Wang, J., Han, X.H., Xu, Y., Lin, L., Hu, H., Jin, C., Chen, Y.W., et~al.: Sparse codebook model of local structures for retrieval of focal liver lesions using multiphase medical images. International journal of biomedical imaging  \textbf{2017} (2017)

\bibitem{wang2022contrastive}
Wang, J., Zeng, Z., Chen, B., Dai, T., Xia, S.T.: Contrastive quantization with code memory for unsupervised image retrieval. In: Proceedings of the AAAI Conference on Artificial Intelligence. vol.~36, pp. 2468--2476 (2022)

\bibitem{wang2017chestx}
Wang, X., Peng, Y., Lu, L., Lu, Z., Bagheri, M., Summers, R.M.: Chestx-ray8: Hospital-scale chest x-ray database and benchmarks on weakly-supervised classification and localization of common thorax diseases. In: Proceedings of the IEEE conference on computer vision and pattern recognition. pp. 2097--2106 (2017)

\bibitem{you2017large}
You, Y., Gitman, I., Ginsburg, B.: Large batch training of convolutional networks. arXiv preprint arXiv:1708.03888  (2017)

\bibitem{barlow}
Zbontar, J., Jing, L., Misra, I., LeCun, Y., Deny, S.: Barlow twins: Self-supervised learning via redundancy reduction (2021)

\bibitem{cb8}
Zhang, Y., Sun, K., Liu, Y., Ou, Z., Shen, D.: Vector quantized multi-modal guidance for alzheimer’s disease diagnosis based on feature imputation. In: International Workshop on Machine Learning in Medical Imaging. pp. 403--412. Springer (2023)

\bibitem{zheng2023online}
Zheng, C., Vedaldi, A.: Online clustered codebook. In: Proceedings of the IEEE/CVF International Conference on Computer Vision. pp. 22798--22807 (2023)

\bibitem{pcrlv2}
Zhou, H.Y., Lu, C., Chen, C., Yang, S., Yu, Y.: A unified visual information preservation framework for self-supervised pre-training in medical image analysis. IEEE Transactions on Pattern Analysis and Machine Intelligence  (2023)

\bibitem{pcrlv1}
Zhou, H.Y., Lu, C., Yang, S., Han, X., Yu, Y.: Preservational learning improves self-supervised medical image models by reconstructing diverse contexts. In: Proceedings of the IEEE/CVF International Conference on Computer Vision. pp. 3499--3509 (2021)

\bibitem{odir1}
Zhou, Y., Wang, B., Huang, L., Cui, S., Shao, L.: A benchmark for studying diabetic retinopathy: segmentation, grading, and transferability. IEEE Transactions on Medical Imaging  \textbf{40}(3),  818--828 (2020)

\end{thebibliography}
\end{document}


\title{CoBooM: Codebook Guided Bootstrapping for Medical Image  Representation Learning}
\subtitle{Supplementary Material}

\author{Anonymous }
\institute{Anonymous}
\maketitle  
